\documentclass{ecai} 

\usepackage{latexsym}
\usepackage{amssymb}
\usepackage{amsmath}
\usepackage{amsthm}
\usepackage{booktabs}
\usepackage{enumitem}
\usepackage{graphicx}
\usepackage{color}
\usepackage{multirow}
\usepackage{caption}
\pdfoutput=1 

\begin{document}


\begin{frontmatter}




\title{Hybrid Feature Collaborative Reconstruction Network \\
for Few-Shot Fine-Grained Image Classification}




\author[A]{\fnms{Shulei}~\snm{Qiu}}
\author[B]{\fnms{Wanqi}~\snm{Yang}}
\author[C]{\fnms{Ming}~\snm{Yang}\thanks{Corresponding Author. Email: myang@njnu.edu.cn.}}

\address[A, B, C]{Nanjing Normal University, China}


\begin{abstract}
Our research focuses on few-shot fine-grained image classification, which faces two major challenges: appearance similarity of fine-grained objects and limited number of samples. To preserve the appearance details of images, traditional feature reconstruction networks usually enhance the representation ability of key features by spatial feature reconstruction and minimizing the reconstruction error. However, we find that relying solely on a single type of feature is insufficient for accurately capturing inter-class differences of fine-grained objects in scenarios with limited samples. In contrast, the introduction of channel features provides additional information dimensions, aiding in better understanding and distinguishing the inter-class differences of fine-grained objects. Therefore, in this paper, we design a new Hybrid Feature Collaborative Reconstruction Network (HFCR-Net) for few-shot fine-grained image classification, which includes a Hybrid Feature Fusion Process (HFFP) and a Hybrid Feature Reconstruction Process (HFRP). In HFRP, we fuse the channel features and the spatial features. Through dynamic weight adjustment, we aggregate the spatial dependencies between arbitrary two positions and the correlations between different channels of each image to increase the inter-class differences. Additionally, we introduce the reconstruction of channel dimension in HFRP. Through the collaborative reconstruction of channel dimension and spatial dimension, the inter-class differences are further increased in the process of support-to-query reconstruction, while the intra-class differences are reduced in the process of query-to-support reconstruction. These designs help the model explore discriminative features of fine-grained objects more accurately, which is crucial for few-shot fine-grained image classification. Ultimately, our extensive experiments on three widely used fine-grained datasets demonstrate the effectiveness and superiority of our approach.
\end{abstract}

\end{frontmatter}


\section{Introduction}
Fine-grained image classification is a long-standing fundamental problem in computer vision \cite{PCM}, which aims to identify fine-grained objects that belong to the same large class but have subtle differences (e.g., bird species, dog species, or car species). In recent years, deep learning has achieved promising results in this field, but it is worth noting that these works \cite{SpdaCNN, RACNN, CAP, APCNN} usually require a large number of labeled samples and strong supervision information such as bounding boxes. In contrast, humans can learn new fine-grained concepts from a few samples. Therefore, in this paper, we study few-shot fine-grained image classification, which is a more challenging task. We try to learn classifiers for new fine-grained classes from a few examples, usually 1 or 5, to mimic human learning abilities.

\begin{figure}[t]
    \includegraphics[width=1\linewidth]{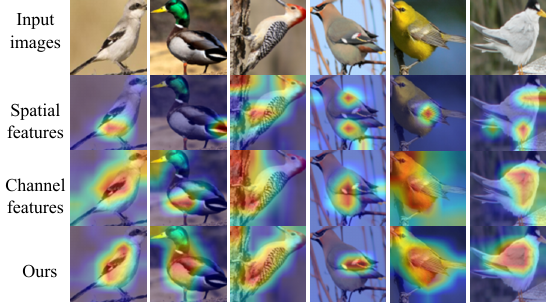}
    \setlength{\belowcaptionskip}{0.3cm}
    \caption{Visual comparison of reconstructions using different features. Compared with reconstruction using spatial or channel features alone, reconstruction using hybrid features (Ours) can provide a more accurate and rich feature representation.}   
  \label{grad-cam}
\end{figure}

The challenges faced in few-shot fine-grained image classification primarily stem from two aspects. Firstly, the fine-grained objects exhibit high visual similarity, making the classification task more difficult. Secondly, the limited number of samples available per fine-grained category makes it challenging for the model to learn comprehensive feature representations from such scarce data. To alleviate these two challenges, researchers have explored various approaches. Metric-based methods \cite{matchingNet, ProtoNet, RealtionNet} aim to address the few-shot learning problem by learning similarity or distance metrics between samples. Due to its effectiveness, researchers have attempted to transfer it from general datasets to fine-grained datasets.

Li et al. \cite{BSNet} learned feature maps based on two similarity measures, enhancing the model's ability to capture discriminative fine-grained features. Huang et al. \cite{LRPABN} introduced a novel low-rank pairwise bilinear pooling operation to effectively capture the subtle differences between support and query images. Wertheimer et al. \cite{FRN} first proposed the reconstruction-based method FRN, which aligns support-query feature pairs by reconstructing query features as weighted sums of support features for each class using ridge regression. This reconstruction-based approach has shown remarkable performance in few-shot fine-grained image classification and has brought new insights to the field. To address the limitation of FRN in fully considering the local features of images during feature reconstruction, Li et al. \cite{LCCRN} introduced a local content extraction module to learn discriminative local features of the image. Furthermore, Wu et al. \cite{BiFRN} proposed a bidirectional feature reconstruction mechanism to overcome the limitation of FRN's unidirectional reconstruction, which only adapts inter-class variation. This method can adapt to both inter-class and intra-class variation, providing new enlightenment for us.

However, these reconstruction-based methods typically rely solely on spatial features, focusing more on discovering the differences in local details among different fine-grained objects, while ignoring the differences in abstract information such as color and texture described by channel features (the recovered images based on channel features in Figure  \ref{recon-visualization} are more significant in color and brightness). When local details are similar, additional supplementary information is crucial for accurately distinguishing fine-grained objects. For instance, in Figure \ref{grad-cam}, we observe that spatial features typically capture tiny spatial regions of the image objects, which may lack sufficient discriminative power.

Therefore, solely relying on spatial features may fail to adequately express the crucial information in the image, and a deeper utilization of channel features may provide a more comprehensive and accurate information foundation for few-shot fine-grained image classification. Although Lee et al. \cite{TDM} proposed weighting each class's channel to locate discriminative regions, we believe this approach simply generates weights based on given support and query sets, without fully considering the differences and importance of channel features, thus, it may still suffer from insufficient channel matching.

To enhance the accuracy and robustness of few-shot fine-grained image classification, we fully exploit the correlation between spatial information and channel information by dynamically generating weights to help the model understand crucial information in the image. Additionally, we introduce channel-wise feature reconstruction to better match and align channel features, enhancing the model's ability to perceive the inter-class differences of fine-grained objects. As shown in the last row of Figure  \ref{grad-cam}, through collaborative feature reconstruction, we can capture both the subtle structures in the image using spatial features and focus on a broader range of features using channel features, thereby overcoming the limitations of using spatial or channel features alone and more comprehensively and accurately capturing the differences among fine-grained object categories.

Our method, the Hybrid Feature Collaborative Reconstruction Network (HFCR-Net), consists of two core components: the hybrid feature fusion process (HFFP) and the hybrid feature reconstruction process (HFRP). Through dynamic weight adjustment, HFFP cleverly captures the spatial dependence between any two positions and accurately aggregates the correlation between channels. Through the update of feature similarity, the optimization and fusion of each feature map are realized, so as to increase the inter-class differences. In the HFRP, we fuse the information of support features and query features spatial relevance and channel relevance by designing four reconstruction tasks. In the process of support-to-query reconstruction, the inter-class differences are further increased, while the intra-class differences are reduced in the process of query-to-support reconstruction. Through the learning of these two processes, we can more accurately explore the discriminative features of fine-grained objects. The ablation experiments in Section 5 show that both processes are indispensable. In addition, we also find that when the spatial feature optimization is performed in parallel with the support feature optimization in HFFP, the network achieves more efficient information learning. The experiments in Section 5 show that this design of parallel construction achieves the best performance.

To sum up, our contributions are three-fold:
\vspace{-5pt} 
\begin{itemize}
    \item We reveal that existing reconstruction-based methods suffer from inaccurate representation of discriminative features due to relying only on spatial features, and then introduced channel features to capture richer inter-class differentiation information.
    \item We propose a new Hybrid Feature Collaborative Reconstruction Network (HFCR-Net) for few-shot fine-grained image classification.
    \item Our extensive experimental results on three fine-grained datasets show that our method outperforms the state-of-the-art methods.
\end{itemize}


\begin{figure*}[ht]
    \centering
    \includegraphics[width=0.95\linewidth]{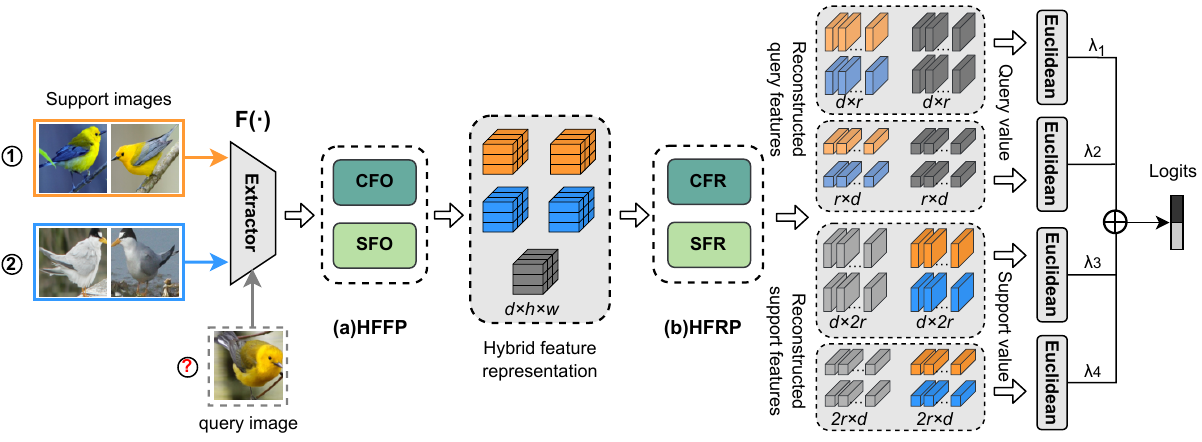}
    \setlength{\abovecaptionskip}{0.2cm}
    \setlength{\belowcaptionskip}{0.3cm}
    \caption{The overview of HFCR-Net. Our network consists of (a) a hybrid feature fusion process (HFFP) and (b) a hybrid feature reconstruction process (HFRP). Among them, HFFP includes channel feature optimization (CFO) and spatial feature optimization (SFO), and HFRP includes channel feature reconstruction (CFR) and spatial feature reconstruction (SFR). In the inference process, after the original image undergoes feature extraction by the feature extractor $F(\cdot)$ to obtain the feature map, we utilize HFFP to aggregate the spatial and channel correlation information of each image and generate the hybrid feature representation. Then, HFRP reconstructs the support features and query features in the channel dimension and spatial dimension respectively, yielding four groups of reconstructed features: channel-reconstructed query features, spatial-reconstructed query features, channel-reconstructed support features, and spatial-reconstructed support features. Finally, we classify the query image into the category with the minimum weighted reconstruction error. $\lambda_{1}$, $\lambda_{2}$, $\lambda_{3}$, and $\lambda_{4}$ denote the corresponding weights, which are used to calculate the overall reconstruction error.}
  \label{network}
\end{figure*}

\section{Related Work}

\textbf{Metric-Based Few-Shot Learning:}
Metric-based few-shot learning methods learn by measuring the similarity between samples. These methods usually focus on how to measure and express the distance or similarity between samples more effectively in the case of a few samples. By learning the prototypes of each category (the central representation in the feature space), ProtoNet \cite{ProtoNet} utilizes Euclidean distance to assign test samples to the nearest prototype. RelationNet \cite{RealtionNet} measures the relationship between pairs of input samples by learning a mapping instead of fixed metric functions such as cosine or Euclidean distance. DN4 \cite{dn4} introduces deep local features into few-shot learning to achieve image-to-class metric representation.

In addition to these classical methods, some metric-based few-shot fine-grained image classification methods have also emerged in recent years. BSNet \cite{BSNet} learns feature maps based on two similarity measures of different features so that the model learns more discriminative and less similarity-biased features from a few fine-grained images. The low-rank pairwise bilinear pooling operation in LRPABN \cite{LRPABN} reveals fine-grained relationships between different support-query image pairs. Different from previous methods, our method measures the similarity between sample pairs by the error before and after feature reconstruction.

\textbf{Feature Reconstruction:}
Feature reconstruction refers to the creation of new feature representations by recombining or transforming the original features. The existing few-shot fine-grained methods based on feature reconstruction can align the spatial positions of different fine-grained image objects to enhance the adaptability of few-shot fine-grained image classification tasks. DeepEMD \cite{deepemd} formulates the reconstruction as an optimal matching problem, which represents the similarity between two images in terms of their optimal matching cost. FRN \cite{FRN} reconstructs feature maps of query samples by exploiting closed-form solutions of ridge regression. BiFRN \cite{BiFRN} introduces support-to-query and query-to-support bidirectional reconstruction mechanisms, which can adapt to both inter-class and intra-class variation. LCCRN \cite{LCCRN} uses global features and local features for cross-reconstruction to focus on the appearance details and local information of the image at the same time. 

Compared with these methods that simply use basic spatial features for reconstruction, our method pays more attention to the comprehensive consideration of channel and spatial dimensions in feature reconstruction and constructs four reconstruction tasks to learn more discriminative and representational feature representations.

\textbf{Hybrid Features in Vision Tasks:}
 Hybrid features usually refer to feature representations that combine features from different types or sources. This feature representation can include information from multiple modalities (such as images, text, etc.) \cite{yang1, yang2, yang3}, or different levels (such as low-level features and high-level features) \cite{ding1, ding2}. In this paper, when we refer to hybrid features, we refer to feature representations that utilize both spatial and channel features. In the field of computer vision, images are usually composed of spatial dimension and channel dimension. Spatial features usually involve the position relationship between pixels in the image, while channel features involve the color, texture, brightness, and other features of the image. Hybrid feature representation can combine spatial information and channel information simultaneously to improve the model's ability to represent complex structures and patterns of images. This method has shown good performance in a variety of vision tasks such as object detection \cite{SCAR, CBAM}, segmentation \cite{DAnet}, and video \cite{video}.

Recently, we have seen successful cases of applying channel features to few-shot fine-grained image classification, such as TDM \cite{TDM}. TDM weights the channels of each class based on channel average pooling to locate discriminative regions of fine-grained image objects. We are also inspired by these examples, but our method is different from theirs. We capture the mutual dependencies between different spatial positions and different channels by dynamically adjusting weights, and we utilize the hybrid feature representation for collaborative reconstruction of spatial and channel dimensions. Experiments show that our model achieves more significant advantages in performance.

In addition, we have recently focused on issues for sample selection in open environments. Yang et al. \cite{yang4} introduce PAL, an efficient sampling scheme that progressively identifies valuable OOD instances by assessing their informativeness and representativeness, thus balancing pseudo-ID and pseudo-OOD instances. Furthermore, Yang et al. \cite{yang5} also present WiseOpen, an OSSL framework that uses a gradient-variance-based selection mechanism to enhance ID classification by selectively leveraging a favorable subset of open-set data for training.
 In the future, we will also further integrate deep research by wisely leveraging open-set data.


\section{Method}

In this part, we first introduce the problem definition of the few-shot fine-grained image classification task and then introduce the framework of our new Hybrid Feature Collaborative Reconstruction Network (HFCR-Net) and the two important processes of the network.

\subsection{Problem Definition}
Given a dataset $D=\left\{\left(x_{i}, y_{i}\right), y_{i} \in Y\right\}$, we get two parts: The meta-training set $D_{base}=\left\{\left(x_{i}, y_{i}\right), y_{i} \in Y_{base}\right\}$ and the meta-testing set $D_{novel}=\left\{\left(x_{i}, y_{i}\right), y_{i} \in Y_{novel}\right\}$, where $x_{i}$ and $y_{i}$ are the original feature vector and class label of the $i^{t h}$ image, respectively. $Y_{base}$ and $Y_{novel}$ denote the base class and the novel class, respectively, and the two parts are disjoint ($Y_{base} \cap Y_{novel}=\emptyset$). For an $N$-way $K$-shot few-shot classification task, training and testing typically consist of episodes, where each episode contains $N$ randomly sampled classes, and each class consists of $K$ labeled examples and $U$ unlabeled examples. A collection of samples of these labels is called the support set $S=\left\{\left(x_{j},y_{j}\right)\right\}_{j=1}^{N\times K}$, collection of samples without a label called query set $Q=\left\{\left(x_{j},y_{j}\right)\right\}_{j=1}^{N\times U}$, these two sets are mutually disjoint ($S\cap Q=\emptyset$). The support and query sets are used for learning and testing, respectively.


\subsection{Framework}
In Figure \ref{network}, we present the overall framework structure of the proposed method. After obtaining the input support-query image pairs, the corresponding feature map is first extracted through the feature encoder $F(\cdot)$. HFFP dynamically adjusts the weight in the feature space, optimizes the channel and spatial dimensions of the feature map of each image, and fuses the optimized features to form a hybrid feature representation with dynamic selection and emphasis on key regions.

In HFRP, we design four reconstruction tasks: channel support features reconstruct channel query features, spatial support features reconstruct spatial query features, channel query features reconstruct channel support features, and spatial query features reconstruct spatial support features. Finally, the Euclidean distance was used to measure the reconstruction error between the original feature and the reconstructed feature, and the weighted sum of all reconstruction errors was used as the basis for classifying query images.


\begin{figure}[t]
  \centering
    \includegraphics[width=1.1\linewidth]{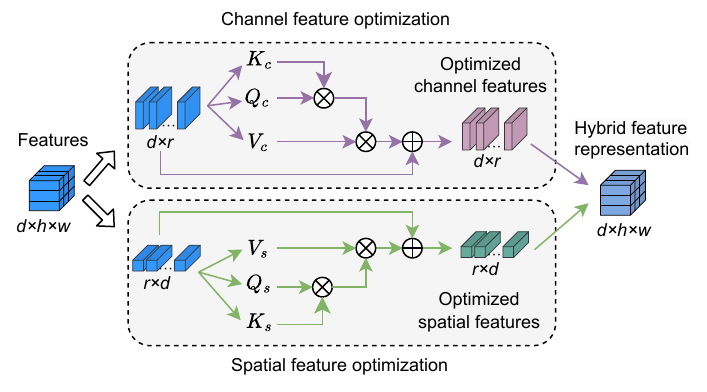}   
    \setlength{\belowcaptionskip}{0.6cm}
    \caption{Hybrid feature fusion process.}
  \label{HFFP}
\end{figure}

\subsection{Hybrid Feature Fusion Process (HFFP)}
HFFP aims to adaptively adjust the channel weight and spatial weight so that the model can better focus on the key image information parts that affect the classification results, and assist the following HFRP at the same time. The HFFP we designed is shown in Figure \ref{HFFP}.

For an $N$-way $K$-shot few-shot classification task, we input $N\times(\mathrm{K+U})$ images into the feature extractor $F(\cdot)$ to extract the corresponding feature maps. For each image $x_{i}$, the feature map is $f_{i}=F(x_{i})\in\mathbb{R}^{d\times h\times w}$, where $d$ is the number of channels and $h$ and $w$ are the height and width of the features, respectively. We implement the optimal fusion process of hybrid features by parallel two feature optimization operations.

\textbf{For Channel Feature Optimization (CFO)},
 we view the feature map $f$ as a sequence of $d$ vectors, each vector contains $r$ elements, where $r=h\times w$, then the feature map can be expressed as $f=[f^{1},f^{2},...,f^{d}]$, where $f^{i}\in \mathbb{R}^{r}$. We multiply each feature map $f_{i}$ with the learnable weights $W_{c}^{Q}$, $W_{c}^{K}$, $W_{c}^{V}$ to obtain $Q_c$, $K_c$, $V_c$, where $W_{c}^{Q}, W_{c}^{K}, W_{c}^{V}\in \mathbb{R}^{r\times r}$. It should be noted that we embed sinusoidal position encoding \cite{attention} in each feature map $f_{i}$. We then use the following formula to compute the channel optimization map $f_{i,c}$:
\begin{flalign}
\begin{split}
    f_{i,c}=\mathrm{Softmax}\left(\frac{Q_{c}K_{c}^{T}}{\sqrt{r}}\right)V_{c} ,f_{i,c}\in \mathbb{R}^{d\times r},
\end{split}
\end{flalign}
where $Q_cK_c^T$ represents the attention score matrix obtained by matrix multiplication and our scaling factor is set to $\frac1{\sqrt{r}}$ to control the size of attention weights.

\textbf{For Spatial Feature Optimization (SFO)}, we view the original feature map $f$ as a sequence of $r$ vectors, each vector containing $d$ elements, then the feature map can be expressed as $f=[f^{1},f^{2},...,f^{r}]$, where $f^{i}\in \mathbb{R}^{d}$. Similarly, the space optimization mapping $f_{i,s}$ can be obtained by the following formula:
\begin{flalign}
\begin{split}
    f_{i,s}=\mathrm{Softmax}\left(\frac{Q_{s}K_{s}^{T}}{\sqrt{d}}\right)V_{s} ,f_{i,s}\in \mathbb{R}^{r\times d},
\end{split}
\end{flalign}
where $Q_s$, $K_s$, and $V_s$ are obtained by matrix multiplication of the original feature map $f_{i}$ with the learnable weights $W_{s}^{Q}$, $W_{s}^{K}$ and $W_{s}^{V}$ respectively, and $W_{s}^{Q}, W_{s}^{K}$, $W_{s}^{V}\in \mathbb{R}^{d\times d}$.

Finally, the channel optimization map $f_{i,c}\in \mathbb{R}^{d\times r}$ and the space optimization map $f_{i,s}\in \mathbb{R}^{r\times d}$ are reshaped into the same dimension as the original feature map $f\in \mathbb{R}^{d\times h\times w}$ and fused to obtain the hybrid feature representation $g_{i}$:
\begin{flalign}
\begin{split}
    g_{i}=f_{i,c}+f_{i,s} , g_{i}\in \mathbb{R}^{d\times h\times w}.
\end{split}
\end{flalign}

\begin{table*}[ht]
\caption{5-way few-shot classification performance of different methods on CUB, Dogs, and Cars datasets under different backbone.}
\renewcommand{\arraystretch}{1.20}
\centering
\begin{tabular}{@{}cccccccc@{}}
\toprule[1.2pt]

\multirow{2}{*}{Backbone} & \multirow{2}{*}{Method} & \multicolumn{2}{c}{CUB}          & \multicolumn{2}{c}{Dogs}    & \multicolumn{2}{c}{Cars}    \\
                          &                         & 1-shot          & 5-shot         & 1-shot       & 5-shot       & 1-shot       & 5-shot       \\ \midrule
\multirow{13}{*}{Conv-4}  & ProtoNet(NeurIPS 2017)* & 64.67±0.23      & 85.67±0.14     & 46.27±0.21   & 70.45±0.16   & 50.74±0.22   & 74.78±0.17   \\
                          & Relation(CVPR 2018)     & 63.94±0.92      & 77.87±0.64     & 47.35±0.88   & 66.20±0.74   & 46.04±0.91   & 68.52±0.78   \\
                          & DN4(CVPR   2019)        & 57.45±0.89      & 84.41±0.58     & 39.08±0.76   & 69.81±0.69   & 34.12±0.68   & 87.47±0.47   \\
                          & DeepEMD(CVPR   2020)    & 64.08±0.50      & 80.55±0.71     & 46.73±0.49   & 65.74±0.63   & 61.63±0.27   & 72.95±0.38   \\
                          & LRPABN(TMM   2021)      & 63.63±0.77      & 76.06±0.58     & 45.72±0.75   & 60.94±0.66   & 60.28±0.76   & 73.29±0.58   \\
                          & BSNet(D\&C)(TIP   2021) & 62.84±0.95      & 85.39±0.56     & 43.42±0.86   & 71.90±0.68   & 40.89±0.77   & 86.88±0.50   \\
                          & FRN(CVPR   2021)*       & 75.44±0.21      & 89.87±0.12     & 60.27±0.22   & 79.17±0.14   & 66.09±0.22   & 86.77±0.12   \\
                          & FRN+TDM(CVPR   2022)*   & 77.41±0.21      & 90.70±0.11     & 62.77±0.22   & 79.71±0.14   & 72.26±0.21   & 89.55±0.10   \\
                          & BiFRN(AAAI   2023)*     & 78.66±0.20      & 92.00±0.11     & 64.48±0.22   & 81.07±0.14   & 74.90±0.20   & \textbf{90.63±0.10}   \\
                          & BiFRN+TDM(CVPR   2022)* & 77.79±0.20      & 91.72±0.11     & 64.00±0.22   & 81.10±0.14   & 75.05±0.20   & 90.58±0.10   \\
                          & LCCRN(TCSVT 2023)*      & 77.83±0.21      & 89.94±0.12     & 64.30±0.22   & 80.36±0.14   & 74.07±0.20  & 89.95±0.10   \\
                          & \textbf{Ours}           & \textbf{80.44±0.20}     & \textbf{92.59±0.11}    & \textbf{66.22±0.22}   & \textbf{81.78±0.14}   & \textbf{75.82±0.20}   & 90.19±0.10   \\ \midrule
\multirow{8}{*}{ResNet-12}& ProtoNet(NeurIPS 2017)* & 69.90±0.20      & 90.65±0.11     & 73.00±0.22   & 86.47±0.13   & 84.56±0.20   & 93.36±0.10   \\
                           & DeepEMD(CVPR   2020)*   & 75.59±0.30      & 88.23±0.18     & 70.38±0.30   & 85.24±0.18   & 80.62±0.26   & 92.63±0.13   \\
                           & FRN(CVPR   2021)*       & 83.16±0.19      & 92.42±0.11     & 75.93±0.22   & 88.72±0.13   & 86.82±0.18   & 94.77±0.09   \\
                           & FRN+TDM(CVPR   2022)*   & 83.26±0.20      & 92.80±0.11     & 75.98±0.22   & 88.70±0.13   & 86.91±0.17   & 96.11±0.07   \\
                           & BiFRN(AAAI   2023)*     & 82.13±0.20      & 93.12±0.11     & 75.89±0.22   & 88.60±0.12   & 87.11±0.17   & 96.06±0.07   \\
                           & BiFRN+TDM(CVPR   2022)* & 82.16±0.20      & 93.38±0.10     & 75.15±0.22   & 88.55±0.12   & 86.73±0.17   & \textbf{96.12±0.07}   \\
                           & LCCRN(TCSVT 2023)*      & 82.38±0.20      & 93.11±0.10     & 75.32±0.22   & 88.33±0.12   & 85.76±0.18   & 96.01±0.07   \\
                           & \textbf{Ours}           & \textbf{84.39±0.19}      & \textbf{93.40±0.11}     & \textbf{77.01±0.22}   & \textbf{88.85±0.13}   & \textbf{87.40±0.17}   & 95.88±0.09
                         \\ \bottomrule[1.2pt]
\end{tabular}
\label{table1} 
\vspace{5pt}  
\end{table*}

\begin{table*}[ht]
\centering
\caption{Ablation experiments using only HFFP or HFRP.}
\begin{tabular}{@{}ccccccccc@{}}
\toprule[1.2pt]
\multirow{2}{*}{Backbone}  & \multirow{2}{*}{HFFP} & \multirow{2}{*}{HFRP} & \multicolumn{2}{c}{CUB}          & \multicolumn{2}{c}{Dogs}    & \multicolumn{2}{c}{Cars}    \\
                           &                       &                       & 1-shot          & 5-shot         & 1-shot       & 5-shot       & 1-shot       & 5-shot \\ \midrule
\multirow{3}{*}{Conv-4}    & \texttimes            & \texttimes            & 64.80±0.23      & 85.92±0.14     & 46.56±0.21   & 71.23±0.16   & 50.88±0.23   & 74.69±0.18   \\
                           & \texttimes            & \checkmark            & 74.94±0.21      & 90.39±0.12     & 61.19±0.21   & 79.88±0.14   & 70.45±0.21   & 88.44±0.11   \\
                           & \checkmark            & \checkmark            & \textbf{80.44±0.20}      & \textbf{92.59±0.11}     & \textbf{66.22±0.22}   & \textbf{81.78±0.14}   & \textbf{75.82±0.20}   & \textbf{90.19±0.10}   \\ \midrule
\multirow{3}{*}{ResNet-12} & \texttimes            & \texttimes            & 81.00±0.20      & 91.90±0.11     & 73.71±0.21   & 87.08±0.12   & 85.86±0.19   & 95.10±0.08   \\
                           & \texttimes            & \checkmark            & 83.59±0.19      & 92.97±0.11     & 74.15±0.22   & 88.09±0.13   & 86.92±0.17   & 95.63±0.07   \\
                           & \checkmark            & \checkmark            & \textbf{84.39±0.19}      & \textbf{93.40±0.11}     & \textbf{77.01±0.22}   & \textbf{88.85±0.13}   & \textbf{87.40±0.17}   & \textbf{95.88±0.09}   \\ \bottomrule[1.2pt]
\end{tabular}
\label{table2} 
\end{table*}

\begin{table*}[ht]
\centering
\caption{Ablation experiments using only channel features or spatial features.}
\begin{tabular}{@{}ccccccccc@{}}
\toprule[1.2pt]
\multirow{2}{*}{Backbone}  & \multirow{2}{*}{SFO + SFR} & \multirow{2}{*}{CFO + CFR} & \multicolumn{2}{c}{CUB}          & \multicolumn{2}{c}{Dogs}    & \multicolumn{2}{c}{Cars}    \\
                           &                           &                          & 1-shot          & 5-shot         & 1-shot       & 5-shot       & 1-shot       & 5-shot \\ \midrule
\multirow{3}{*}{Conv-4}    & \checkmark                & \texttimes               & 77.55±0.20      & 91.10±0.11     & 63.38±0.22   & 79.07±0.14   & 72.90±0.20   & 88.63±0.10   \\
                           & \texttimes                & \checkmark               & 73.57±0.22      & 87.73±0.13     & 60.98±0.23   & 78.29±0.15   & 67.78±0.22   & 84.24±0.14   \\
                           & \checkmark                & \checkmark               & \textbf{80.44±0.20}      & \textbf{92.59±0.11}     & \textbf{66.22±0.22}   & \textbf{81.78±0.14}   & \textbf{75.82±0.20}   & \textbf{90.19±0.10}   \\ \midrule
\multirow{3}{*}{ResNet-12} & \checkmark                & \texttimes               & 80.03±0.20      & 91.32±0.11     & 75.89±0.22   & 88.60±0.12   & 85.11±0.17   & 95.12±0.07   \\
                           & \texttimes                & \checkmark               & 79.86±0.21      & 90.83±0.12     & 74.93±0.22   & 87.55±0.13   & 84.21±0.23   & 95.10±0.16   \\
                           & \checkmark                & \checkmark               & \textbf{84.39±0.19}      & \textbf{93.40±0.11}     & \textbf{77.01±0.22}  & \textbf{88.85±0.13}   & \textbf{87.40±0.17}   & \textbf{95.88±0.09}   \\ \bottomrule[1.2pt]
\end{tabular}
\label{table3} 
\vspace{5pt}  
\end{table*}

\begin{table*}[ht]
\centering
\caption{Effect of the arrangement of CFO and SFO on performance in HFFP.}
\begin{tabular}{@{}cccccccc@{}}
\toprule[1.2pt]
\multirow{2}{*}{Backbone}  & \multirow{2}{*}{Arrangement}    & \multicolumn{2}{c}{CUB}          & \multicolumn{2}{c}{Dogs}    & \multicolumn{2}{c}{Cars}    \\
                           &                                 & 1-shot          & 5-shot         & 1-shot       & 5-shot       & 1-shot       & 5-shot \\ \midrule
\multirow{3}{*}{Conv-4}    & $\mathrm{CFO} \to \mathrm{SFO}$ & 78.94±0.20      & 91.36±0.11     & 65.50±0.22   & 80.90±0.14   & 75.13±0.21   & 89.50±0.11   \\
                           & $\mathrm{SFO} \to \mathrm{CFO}$ & 80.18±0.20      & 92.19±0.11     & 65.97±0.22   & 81.38±0.14   & 74.91±0.21   & 89.69±0.11   \\
                           & CFO \& SFO in Parallel          & \textbf{80.44±0.20}      & \textbf{92.59±0.11}     & \textbf{66.22±0.22}   & \textbf{81.78±0.14}   & \textbf{75.82±0.20}   & \textbf{90.19±0.10}   \\ \midrule
\multirow{3}{*}{ResNet-12} & $\mathrm{CFO} \to \mathrm{SFO}$ & 83.58±0.20      & 92.98±0.11     & 75.89±0.22   & 87.56±0.13   & 86.32±0.18   & 94.82±0.08   \\
                           & $\mathrm{SFO} \to \mathrm{CFO}$ & 82.98±0.20      & 91.34±0.12     & 73.32±0.23   & 85.21±0.14   & 83.49±0.20   & 91.77±0.12   \\
                           & CFO \& SFO in Parallel          & \textbf{84.39±0.19}      & \textbf{93.40±0.11}     & \textbf{77.01±0.22}   & \textbf{88.85±0.13}   & \textbf{87.40±0.17}   & \textbf{95.88±0.09}   \\ \bottomrule[1.2pt] 
\end{tabular}
\label{table4} 
\end{table*}

\subsection{Hybrid Feature Reconstruction Process (HFRP)}
To make full use of the feature information output by HFFP, we design a new hybrid feature reconstruction process (HFRP). For the hybrid feature representation obtained by HFFP, we construct four different reconstruction tasks to reconstruct the original support features and the original query features. Specifically, we divide these four reconstruction tasks into two dimensions: support-query feature reconstruction on the channel dimension and support-query feature reconstruction on the spatial dimension.

\begin{figure}[t]
  \centering
    \includegraphics[width=0.8\linewidth]{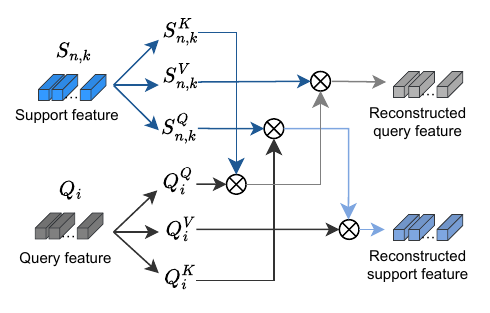}
    \setlength{\belowcaptionskip}{0.6cm}
    \caption{Illustration of feature reconstruction process.}
  \label{HFRP}
\end{figure}

\textbf{For Channel Feature Reconstruction (CFR)}, after HFFP, we can obtain the $k^{th}$ support feature representation of the $n^{th}$ class, i.e. $S_{n,k}=g_{k}^{n}\in \mathbb{R}^{d\times r}$, where $n\in[1,\ldots,N]$,$k\in[1,...,K]$, $r=h\times w$, and the query feature representation $Q_{i}=g_{i}\in \mathbb{R}^{d\times r}$, where $i\in[1,...,N\times U]$. In order to save memory, we make the $S_{n,k}$ and $Q_{i}$ multiplied by the weight of the same parameters, $A_{c}^{Q}$, $A_{c}^{K}$, $A_{c}^{V}$ respectively get $S_{n,k}^{Q}$, $S_{n,k}^{K}$, $S_{n,k}^{K}$ and $Q_{i}^{Q}$, $Q_{i}^{K}$, $Q_{i}^{K}$, Where $A_{c}^{Q} , A_{c}^{K} , A_{c}^{V}\in \mathbb{R}^{r\times r}$. Since the reconstruction process is similar, in Figure \ref{HFRP} we give general examples of $S_{n,k}$ reconstruction $Q_{i}$ and $Q_{i}$ reconstruction $S_{n,k}$.

Then, computing the $i^{th}$ channel-reconstructed query feature $Q_{i,n}^{c}$ from the support features of the $n^{th}$ class can be expressed as the average reconstruction of the $i^{th}$ query feature from the $k$ support features of the $n^{th}$ class. The following equation can describe such a process:
\begin{flalign}
\begin{split}
    Q_{i,n}^{c}=\frac{1}{K}\sum_{k=1}^{K}\mathrm{Softmax}\left(\frac{Q_{i}^{Q}\left(S_{n,k}^{K}\right)^{T}}{\sqrt{r}}\right)S_{n,k}^{V} ,Q_{i,n}\in \mathbb{R}^{d\times r}.
\end{split}
\end{flalign}

Computing the channel-reconstructed support features $S_{n,i}^{c}$ of the $n^{th}$ class from the $i^{th}$ query feature can be expressed as the concatenate of reconstructing $k$ support features of the $n^{th}$ class from the $i^{th}$ query feature. We describe such a process by the following equation:
\begin{flalign}
\begin{split}
    S_{n,i}^{c}=\mathrm{Concat}\left(\mathrm{Softmax}\left(\frac{S_{n,k}^{Q}(Q_{i}^{K})^{T}}{\sqrt{r}}\right)Q_{i}^{V}\right),S_{n,i}\in \mathbb{R}^{d\times kr},
\end{split}
\end{flalign}
where $k\in\left[1,...,K\right]$. Finally, we carry the features after channel reconstruction and the features before reconstruction into residual connection operation: $\hat{Q}_{i,n}^{c}=Q_{i,n}^{c}+Q_{i},\hat{S}_{n,i}^{c}=S_{n,i}^{c}+S_{n}$, where $S_{n}=[S_{n,k}]\in \mathbb{R}^{d\times kr}$,$k\in[1,...,K]$. Injecting the hybrid feature representation directly into the reconstructed features helps to ensure that the model can retain more of the original information when performing image reconstruction, thereby improving the accuracy of reconstruction while mitigating the risk of information loss.

\textbf{For Spatial Feature Reconstruction (SFR)}, we mainly follow the strategy of Wu et al. \cite{BiFRN}. He reshaped the support feature of $n^{th}$ class as $S_{n}\in \mathbb{R}^{kr\times d}$ and the $i^{th}$ query feature as $Q_{i}\in \mathbb{R}^{r\times d}$. And then multiplied by the weight parameters $A_{s}^{Q}$, $A_{s}^{K}$, $A_{s}^{V}$ respectively get $\widetilde{S}_{n,k}^{Q}$, $\widetilde{S}_{n,k}^{K}$, $\widetilde{S}_{n,k}^{K}$ and $\widetilde{Q}_{i}^{Q}$, $\widetilde{Q}_{i}^{K}$, $\widetilde{Q}_{i}^{K}$, where $A_{s}^{Q} , A_{s}^{K} , A_{s}^{V}\in \mathbb{R}^{d\times d}$. Finally, the attention formula is used to calculate the $i^{th}$ spatial-reconstructed query feature $Q_{i,n}^{s}$ from $S_{n}$ and the spatial-reconstructed support feature $S_{n,i}^{s}$ of the $n^{th}$ class from $Q_{i}$. Similarly, for the spatial feature reconstruction process, we obtain $\hat{Q}_{i,n}^{s}$ and $\hat{S}_{n,i}^{s}$ through residual connection.

\begin{figure*}[ht]
    \centering
    \includegraphics[width=0.9\linewidth]{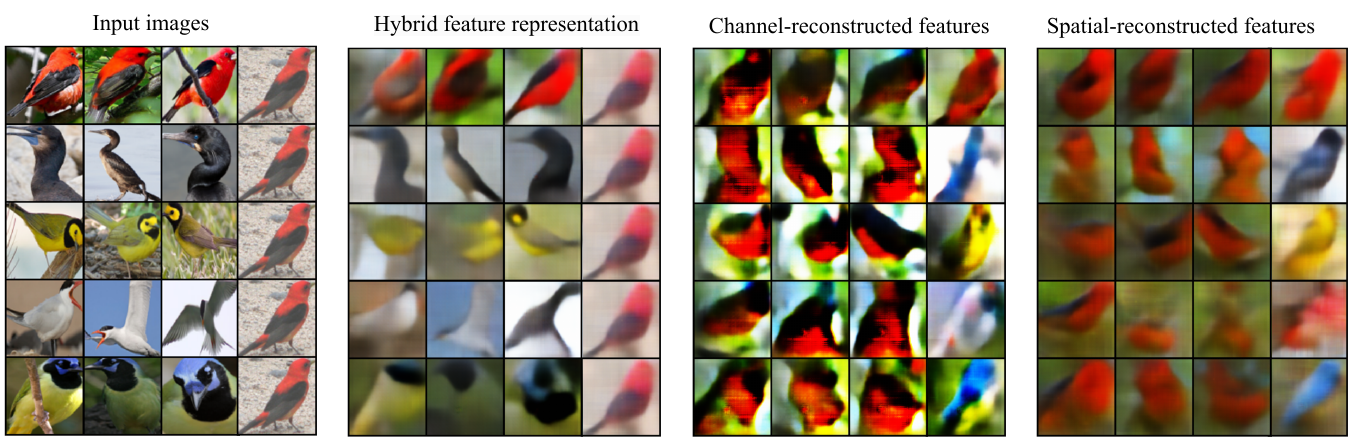}
    \setlength{\abovecaptionskip}{0.3cm}
    \setlength{\belowcaptionskip}{0.3cm}
    \caption{Recovered images using different types of features for the CUB dataset.}
  \label{recon-visualization}
\end{figure*}

\subsection{Objective Function}
After HFRP, for the $i^{th}$ query image $Q_{i}$, we have two kinds of reconstructed feature maps: $\hat{Q}_{i,n}^{c}$ and $\hat{Q}_{i,n}^{s}$. Then the reconstruction error of reconstructing the $i^{th}$ query image from the $n^{th}$ class of support images can be obtained using Euclidean distance:
\begin{flalign}
\begin{split}
    d_{n\to i}^c=\left\|\hat{Q}_{i,n}^c-Q_i^V\right\|^2, d_{n\to i}^s=\left\|\hat{Q}_{i,n}^s-\widetilde{Q}_i^V\right\|^2 .    
\end{split}
\end{flalign}

Similarly, for the $n^{th}$ class of support image $S_{n}$, we also have two kinds of reconstructed feature maps: $\hat{S}_{n,i}^{c}$ and $\hat{S}_{n,i}^{s}$. Then the reconstruction error of the $n^{th}$ class of support image reconstructed from the $i^{th}$ query image are as follows.
\begin{flalign}
\begin{split}
    d_{i\to n}^c=\left\|\hat{S}_{n,i}^c-S_n^V\right\|^2, d_{i\to n}^s=\left\|\hat{S}_{n,i}^s-\widetilde{S}_n^V\right\|^2 ,   
\end{split}
\end{flalign}
where $S_{n}^{V}=[S_{n,k}^{V}]\in \mathbb{R}^{d\times kr}$,$k\in[1,...,K]$. 

The final reconstruction error is the weighted sum of the above four reconstruction errors:
\begin{flalign}
\begin{split}
    d_{(n,i)}=\tau(\lambda_{1}d_{n\to i}^{c}+\lambda_{2}d_{n\to i}^{s}+\lambda_{3}d_{i\to n}^{c}+\lambda_{4}d_{i\to n}^{s}) ,  
\end{split}
\end{flalign}
where $\lambda_{1}$, $\lambda_{2}$, $\lambda_{3}$, and $\lambda_{4}$ are four learnable weight factors, respectively, and the initial value is set to 0.5. $\tau$ is a learnable temperature factor \cite{FRN, BiFRN}.

Based on the final reconstruction error, we compute the probability that the $i^{th}$ query image belongs to the $n^{th}$ class:
\begin{flalign}
\begin{split}
    P(\hat y_i=n|x_i)=\frac{e^{-d_{(n,i)}}}{\sum_{n=1}^{N}e^{-d_{(n,i)}}}  .
\end{split}
\end{flalign}
Then the total loss of the $N$-way $K$-shot few-shot classification task is as follows.
\begin{flalign}
\begin{split}
    L=-\frac{1}{N\times U}\sum_{i=1}^{N\times U}\sum_{n=1}^{N}1(y_{i}==n)\log\left(P(\hat{y}_{i}=n|x_{i})\right)  ,
\end{split}
\end{flalign}
where $1(y_{i}==n)$ means that the formula is equal to 1 if $y_{i}$ and $n$ are equal, and 0 otherwise.

At training time, we update the network by minimizing $L$. At test time, we compute the predicted probability of the query image in Formula (9) for all classes and then classify it to the class with the highest probability.


\section{Experiments}
\subsection{Dataset}
We use three benchmark datasets to evaluate the performance of the proposed method: CUB-200-2011 \cite{CUB}, Stanford Dogs \cite{dogs}, and Stanford Cars \cite{cars}.

\textbf{Cub-200-2011 (CUB)} is a classic fine-grained image classification dataset. It contains 11,788 images from 200 bird species. We crop each image through a manually annotated bounding box \cite{deepemd}. Following \cite{closer}, we split this dataset, and our split is the same as \cite{FRN}.

\textbf{Stanford Dogs (Dogs)} is also a commonly used benchmark dataset for fine-grained image classification, which is a challenging dataset. The dataset contains a total of 20,580 images of 120 dog categories from around the world, covering a wide range of body sizes, colors, and breeds. Our split is the same as \cite{dn4}.

\textbf{Stanford Cars (Cars)} contains 16185 images of 196 classes of cars, each of which corresponds to a different car make and model. Again, we adopt the same data-splitting method as \cite{dn4}.

For the above three datasets, we follow the same evaluation scheme as the previous work \cite{dn4, FRN, BiFRN}, and the images of each dataset are resized to 84×84.

\subsection{Implementation Details}

We conduct experiments with two widely used backbone architectures: Conv-4 and ResNet-12. Identical to recent works\cite{FRN, zhang2021rethinking} on few-shot classification, Conv-4 consists of four identical convolutional blocks concatenated, each with a down-sampling factor of two. For an 84×84 image input, Conv-4 generates feature maps of size 64×5×5 and ResNet-12 generates feature maps of size 640×5×5.

In training, we use standard data augmentation with random cropping, horizontal flipping, and color jitter according to  \cite{closer, wang2019simpleshot, FRN, deepemd} to get better training stability. Our experiments are conducted via Pytorch \cite{pytorch} on an NVIDIA 3090Ti GPU. We set the initial learning rate to 0.1, and the weight decay to 5e-4, and for all models, we trained the model for 1200 epochs using SGD with Nesterov momentum of 0.9, with the learning rate decreasing by a scaling factor of 10 after every 400 epochs. When the backbone architecture is Conv-4, we train the model with 20-way 5-shot episodes and test the model with 5-way 1-shot and 5-shot episodes. When the backbone architecture is ResNet-12, we train the model with 5-way 5-shot episodes to save memory and test the model with 5-way 1-shot and 5-shot episodes. Under both backbone architectures, for each class, we use 15 queries, and we validate the models at every 20 epochs and leave the best-performing model. In addition, for all experiments, we report the average accuracy on the test set for 10,000 randomly generated episodes, where each class contains 16 queries and the confidence interval is set to 95\%.

\subsection{Comparison with Existing Methods}
We conduct experiments on the three benchmark fine-grained datasets discussed earlier to verify the performance of our proposed method on the few-shot fine-grained image classification task. In Table \ref{table1}, we show ten representative methods, including five classical general few-shot classification methods (ProtoNet \cite{ProtoNet}, Relation \cite{RealtionNet}, DN4 \cite{dn4}, DeepEMD \cite{deepemd}, FRN \cite{FRN}) and five specialized few-shot fine-grained classification methods (LRPABN \cite{LRPABN}, BSNet \cite{BSNet}, TDM \cite{TDM}, BiFRN \cite{BiFRN}, LCCRN \cite{LCCRN}). For fairness, we also compare TDM embedding into BiFRN with our method, because TDM locates discriminative regions by weighting the channels of each class. The results of the method labeled * are those that we reproduce under the same experimental Settings as our method according to the open-source code.

We use Conv-4 and ResNet-12 as the backbone of all compared methods. Compared with the above general few-shot classification methods and specialized few-shot fine-grained classification methods, our method achieves the best performance in most cases. Except the performance under the 5-way 5-shot setting on the Cars dataset is slightly lower than the previous methods. Overall, the performance improvement under the 5-way 5-shot setting is not as significant as that under the 5-way 1-shot setting. We guess that this is due to the larger amount of training data in the 5-way 5-shot setting compared to the 5-way 1-shot setting. This means that in the 5-way 5-shot task, the model has more training samples for learning features and patterns, and thus can achieve a decent result even without using mixed feature representations.


\section{Analysis}
On the above three datasets, we use Conv-4 and ResNet-12 as the backbone networks to conduct ablation experiments, respectively, and then study the effectiveness of the design and components of our method.

\subsection{Ablation Experiment}
\textbf{The Effectiveness of HFFP and HFRP: }In Table \ref{table2}, we compare the effect of HFFP or HFRP on the experimental results. We try to remove either HFFP or HFRP or remove both of them, while the rest of the Settings remain unchanged. Deleting both of them is equivalent to ProtoNet \cite{ProtoNet}.

Clearly, removing either or both of these results in worse experiments. This indicates that the hybrid feature representation after HFFP optimization fusion can help the reconstruction accuracy of HFRP, and both are indispensable.

\textbf{The Effectiveness of Using Hybrid Features: }Table \ref{table3} reports the effect of removing channel features or spatial features in our model on the experimental results. We observe that using only spatial features performs better than using only channel features because fine-grained classification involves distinguishing small differences among classes, while spatial features are more sensitive to these geometric structures and details. In the last row of the table, we see a significant improvement when using hybrid features. This confirms that channel features and spatial features are complementary and the use of hybrid features plays a key role in improving model performance.

\textbf{The Method of Combining CFO and SFO in HFFP: }
In  Table \ref{table4}, we compare the arrangement of three different CFO and SFO in HFFP: first CFO then SFO, first SFO then CFO, or use both of them in parallel. Due to the difference in information carried by each type of feature, different permutations may affect the overall performance.

From the results, we find that in the case of backbone Conv-4, using SFO preferentially leads to better performance than using CFO preferentially. However, in the ResNet-12 backbone case, prioritizing CFO leads to better performance than prioritizing SFO. Taken together, parallelizing both of them can further improve performance. Notably, all arrangement strategies outperform using one type of feature alone, which further indicates that the use of hybrid features can effectively exploit the complementarity of channel features and spatial features to provide more comprehensive and rich information, and the parallel arrangement further improves accuracy.

\subsection{Visual Analysis of Reconstruction}
To better illustrate the effectiveness of our proposed method, we visualize the features. We trained an inverted ResNet-12 as the decoder network, for which the input is the feature value and the output is a 3×84×84 restored image. Our decoder network is trained using the Adam optimizer with an initial learning rate set to 0.01, a batch size of 200, and 1000 training times with a 4-fold decrease in the learning rate every 100 epochs.

As shown in Figure \ref{recon-visualization}, the leftmost block is the original support images and the query image, and the support images contain five classes with three images for each class. The query image was replicated 5 times for comparison. The second left block is recovered from the corresponding unreconstructed hybrid feature representation. The third block from the left is the recovered images of the channel-reconstructed features. The last block is the recovered images of the spatial-reconstructed features. The first row represents the same class and the remaining four rows represent different classes.

Obviously, the reconstructed images based on the same class are visually more similar to the original images. The reconstructed images based on different classes are visually very different from the original images. However, the recovery of the channel-based reconstructed images is more significant in color and brightness, while the recovery of the spatial-based reconstructed images is more accurate in contour and position. It is sufficient to show that the hybrid feature-based reconstruction can provide complementary information about the original image.


\section{Conclusion}
In this paper, we propose a novel Hybrid Feature Collaborative Reconstruction Network (HFCR-Net) for few-shot fine-grained image classification. Different from traditional methods, the proposed method not only focuses on the position relationship between pixels but also comprehensively considers the weight allocation of channel features. In the case of limited samples, utilizing hybrid features enables a more comprehensive and accurate capture of the inter-class differences of fine-grained objects. Extensive experiments show that the classification performance of HFCR-Net achieves better results in few-shot fine-grained image classification tasks. Furthermore, we plan to continue our research on few-shot fine-grained image classification by utilizing open-set data in the future.


\bibliography{mybibfile}

\end{document}